\pdfoutput=1

\documentclass[11pt]{article}

\usepackage{acl}

\usepackage{microtype}
\usepackage{hyperref}
\usepackage{url}
\usepackage{booktabs}

\usepackage{times}
\usepackage{latexsym}

\usepackage[T1]{fontenc}

\usepackage[utf8]{inputenc}

\usepackage{enumerate}
\usepackage{graphicx}
\usepackage[skip=0pt]{caption}
\usepackage{graphics, amsmath}
\usepackage{amssymb}
\usepackage{multirow}

\usepackage{tabularx}
\usepackage{pifont}
\usepackage{colortbl}
\usepackage{xspace}
\usepackage{xcolor}
\usepackage[most]{tcolorbox}
\usepackage{makecell}
\usepackage{adjustbox}
\usepackage[inline]{enumitem}

\usepackage{arydshln}
\usepackage{acronym}

\usepackage{wrapfig}

\usepackage{blindtext}
\usepackage{ragged2e}

\definecolor{myblue0}{RGB}{192, 214, 234}
\definecolor{myblue1}{RGB}{166, 199, 226}
\definecolor{mygray}{RGB}{224, 224, 232}

\definecolor{OliveGreen}{rgb}{0,0.6,0}
\definecolor{myred}{HTML}{F8CCCC}
\definecolor{mygreen}{HTML}{E0ECD4}

\newcommand{\fetaqa}{FeTaQA\xspace}
\newcommand{\qtsumm}{QTSumm\xspace}
\newcommand{\qfmts}{QFMTS\xspace}

\newcommand{\directsumm}{DirectSumm\xspace}
\newcommand{\reasonsumm}{Reason-then-Summ\xspace}

\newcommand{\binder}{Binder\xspace}
\newcommand{\dater}{Dater\xspace}
\newcommand{\refactor}{ReFactor\xspace}
\newcommand{\tapera}{TaPERA\xspace}

\newcommand{\qdmr}{QDMR\xspace}

\newcommand{\qpl}{QPL\xspace}

\newcommand{\ourplan}{TaSoF\xspace}
\newcommand{\tabreasonours}{SPaGe\xspace}

\title{Beyond Natural Language Plans: Structure-Aware Planning for Query-Focused Table Summarization}

\author{Weijia Zhang$^1$ \quad Songgaojun Deng$^1$ \quad Evangelos Kanoulas$^1$  \\
$^1$IRLab, University of Amsterdam \\
\texttt{w.zhang2@uva.nl}}

\begin{document}
\maketitle

\begin{abstract}

Query-focused table summarization requires complex reasoning, often approached through step-by-step natural language (NL) plans. However, NL plans are inherently ambiguous and lack structure, limiting their conversion into executable programs like SQL and hindering scalability, especially for multi-table tasks. To address this, we propose a paradigm shift to structured representations. We introduce a new structured plan, \ourplan, inspired by formalism in traditional multi-agent systems, and a framework, \tabreasonours, that formalizes the reasoning process in three phases: 
\begin{enumerate*}[label=\arabic*),nosep]
    \item \textit{Structured Planning} to generate \ourplan from a query
    \item \textit{Graph-based Execution} to convert plan steps into SQL and model dependencies via a directed cyclic graph for parallel execution, and
    \item \textit{Summary Generation} to produce query-focused summaries.
\end{enumerate*}
Our method explicitly captures complex dependencies and improves reliability. Experiments on three public benchmarks show that \tabreasonours consistently outperforms prior models in both single- and multi-table settings, demonstrating the advantages of structured representations for robust and scalable summarization.

\end{abstract}

\begin{figure}[th]
    \centering
    \includegraphics[width=0.995\linewidth]{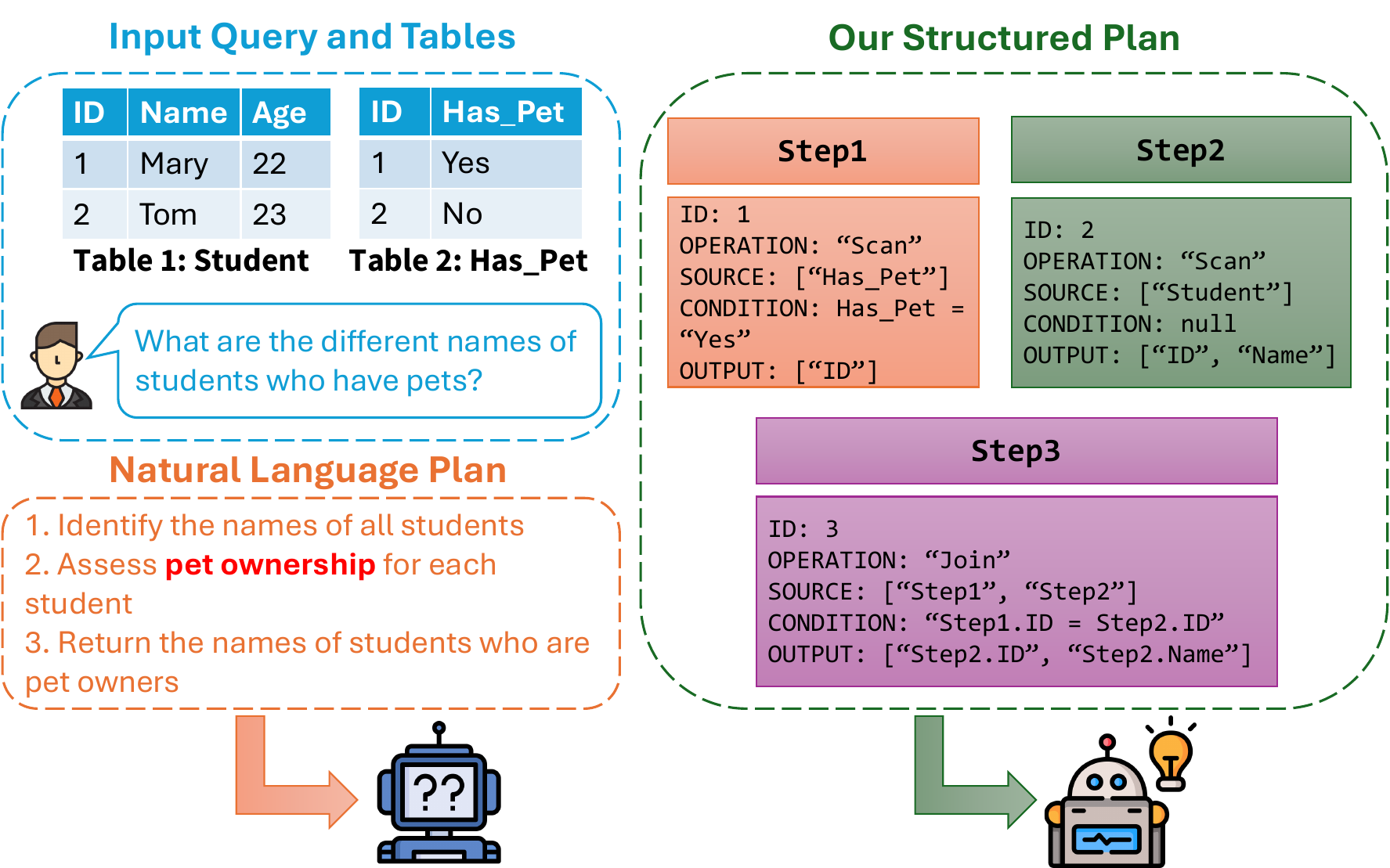}
    \caption{An example comparing a natural language plan with our proposed structured plan, \ourplan. The natural language plan includes vague terms such as \textit{pet ownership}, which can be hard for LLMs to interpret. In contrast, \ourplan follows standard class definitions and specifies table-specific operations, making it easier for LLMs to understand.}
    \label{fig:tabreason:motivation_example}
\end{figure}
\section{Introduction}
\label{sec:tabreason:intro}

Query-focused table summarization aims to generate a concise, textual summary from tabular data, specifically tailored to a user's query. This task has recently gained significant attention, emerging as a critical challenge in the field of table understanding~\cite{zhao-etal-2023-qtsumm,zhao-etal-2024-tapera,zhang2024qfmts}.
A central hurdle in this domain is table reasoning, which involves accurately extracting and aggregating query-relevant information from potentially complex input tables.
Earlier studies predominantly focused on training end-to-end models that performed reasoning implicitly~\cite{liu2022tapex,liu-etal-2022-plog,jiang-etal-2022-omnitab}. While these approaches showed promising results, they often required extensive training data and lacked interpretability~\cite{nguyen2025interpretable}.

With the rapid advancements in large language models (LLMs), recent research has shifted towards explicit, multi-step table reasoning. These cutting-edge methods typically leverage techniques like query decomposition~\cite{ye2023dater,zhao-etal-2024-tapera,nguyen2025interpretable} or pre-defined workflows~\cite{wang2024cotable,zhang-etal-2024-e5,li2024graphotter} to generate a step-by-step reasoning plan and then execute these steps to perform the necessary table operations.

However, a significant limitation persists: these reasoning plans are overwhelmingly generated as free-form natural language (NL). This introduces substantial challenges because NL plans are inherently unreliable and lack the explicit structure crucial for robust execution. This linguistic ambiguity makes the automated conversion of plan steps into executable programs for table operations, such as SQL queries, a brittle and error-prone process; even minor variations in phrasing can lead to critical execution failures~\cite{zhao-etal-2024-tapera,nguyen2025interpretable}. Furthermore, the absence of a formal, machine-readable structure severely limits the scalability of these methods to complex multi-table scenarios and hinders opportunities for crucial execution optimizations.

To address these limitations, we propose a paradigm shift from ambiguous NL plans to a formal, structured representation. 
Inspired by formalism in traditional multi-agent systems~\cite{dinverno1997formalism,hilaire2000formal,gruer2002formal}, we standardize the reasoning plan by defining it as a \underline{Ta}ble \underline{S}tate and \underline{o}peration \underline{F}low (\ourplan). 
In this framework, each planning step is defined as a \textit{Step} class with a schema. Specifically, the \textit{Step} class defines a specific table operation with its input and output states. 
An example of comparison between an NL plan and \ourplan is shown in \autoref{fig:tabreason:motivation_example}.
From this example, we observe that natural language plans often include linguistic ambiguities, such as the phrase ``\textit{pet ownership}'' in the second step, which can be difficult for the LLM-based execution module to interpret accurately. In contrast, our structured plan adopts class definitions from software engineering, using explicit expressions like \textit{Has\_Pet = ``Yes''}, which are easier to understand and translate into executable programs.
Built upon this, we then present \tabreasonours (\underline{S}tructured \underline{P}lanning \underline{a}nd \underline{G}raph-based \underline{e}xecution), a new framework designed to robustly handle both single- and multi-table summarization scenarios. Our framework operates in three distinct phases:
\begin{enumerate*}[label=\arabic*),nosep]
\item \textit{Structured Planning}: Given a complex user query, an LLM-based planner generates a hierarchical, multi-step structured plan in the \ourplan format. 
This key-value structure explicitly models key cross-step dependencies and is more concise and less ambiguous than its natural language counterpart.
\item \textit{Graph-based Execution}: For each step within the structured plan, an executor leverages the explicit key-value structure to convert the step into a reliable executable SQL query. An external code interpreter then executes this query, yielding the intermediate output table. This structured representation enables us to model the entire plan as a Directed Acyclic Graph (DAG), facilitating the efficient parallel execution of independent steps.
\item \textit{Summary Generation}: Once all plan steps have been successfully executed and intermediate results are obtained, a dedicated summary generator produces a human-readable natural language summary tailored to the user query.
\end{enumerate*}

We thoroughly evaluate \tabreasonours against existing methods on three public query-focused table summarization datasets: \fetaqa, \qtsumm, and \qfmts. Our experimental results consistently demonstrate that our framework significantly outperforms current baseline models, underscoring its effectiveness. Extensive analysis further reveals the superiority of our proposed structured plan format compared to natural language plans, leading to more robust and reliable execution. We also provide qualitative analysis to illustrate how our structured plan enables smoother and more accurate execution.

To the best of our knowledge, our method is the first to formalize ambiguous natural language plans into a fully structured, machine-readable format specifically for query-focused table summarization.

Our key contributions are summarized as follows:
\begin{itemize}[leftmargin=12pt,topsep=2pt,itemsep=-2pt]%
    \item We define \ourplan, a novel structure that formalizes reasoning plans as a flow of table states and operations, thereby improving execution efficiency and accuracy.
    \item We propose \tabreasonours, a new structure-aware framework that leverages this structured planning and graph-based execution to achieve superior performance on the query-focused table summarization task.
    \item Extensive experiments and analysis on three benchmarks demonstrate the effectiveness of our approach in complex multi-table reasoning and its ability to enable efficient parallel execution.
\end{itemize}

\section{Related Work}
\label{sec:tabreason:related}

This section reviews two key areas of related work: program-augmented table reasoning and planning for table understanding.

\subsection{Program-Augmented Table Reasoning}

Table reasoning necessitates models that can mimic human data analysis, encompassing operations like arithmetic calculations (e.g., counting, aggregation) and the judicious filtering of irrelevant rows or columns. This capability is a fundamental component for various table understanding tasks. Earlier approaches~\cite{liu2022tapex,jiang-etal-2022-omnitab} primarily relied on end-to-end reasoning, fine-tuning black-box models to directly produce task-specific outputs. A significant limitation of these methods is their inherent lack of interpretability, making it challenging to understand the intricate intermediate reasoning steps.
More recent studies~\cite{cheng2023binder,ye2023dater,zhang2024reactable} have adopted program-augmented reasoning frameworks, often by employing tool-augmented LLMs~\cite{schick2023toolformer}. These approaches generate executable, task-oriented code, typically SQL or Python, to derive intermediate sub-tables or results. This paradigm significantly enhances interpretability by making the explicit reasoning steps traceable and verifiable. 
However, most existing program-augmented methods primarily target single-table scenarios, which restricts their direct applicability to complex real-world tasks that inherently demand reasoning across multiple, interconnected tables. 
Our work explicitly addresses both single- and multi-table scenarios, and crucially, introduces a more efficient graph-based execution paradigm to further optimize this process.

\subsection{Planning for Table Understanding}

The remarkable advancements in large language models (LLMs) have spurred a surge of recent studies exploring planning mechanisms for various table understanding tasks, including table summarization~\cite{zhao-etal-2023-qtsumm,zhang2024qfmts}, table question answering~\cite{pasupat-liang-2015-compositional,nan-etal-2022-fetaqa}, and table fact verification~\cite{chen2020tabfact}.
Existing LLM-based planning approaches in this domain can be broadly categorized into two main paradigms:
\begin{enumerate*}[label=\arabic*),nosep]
\item \textit{Query Decomposition}: This paradigm involves prompting LLMs to decompose a complex, original query into a series of smaller, more manageable sub-queries. These sub-queries are then addressed sequentially, with their intermediate results contributing to the final answer for the overall query~\cite{ye2023dater,zhao-etal-2024-tapera}.
\item \textit{Pre-defined Workflow}: In this paradigm, LLMs are guided to iteratively select and execute operations from a pre-defined set or to follow a sequence of human-engineered stages. This process continues until a specific stop condition is met, leading to the desired outcome~\cite{wang2024cotable,zhang-etal-2024-e5,li2024graphotter,mao2024potable}.
\end{enumerate*}
While these methods have demonstrated effectiveness, particularly in single-table scenarios, they predominantly rely on natural language (NL) plans. A critical limitation of NL plans is their inherent linguistic ambiguity, which can hinder robust and reliable subsequent execution processes. In stark contrast, our work is the first to explicitly explore and formalize structure-based planning for both single- and multi-table scenarios, addressing the limitations of NL-centric approaches.

\begin{figure}[th]
    \centering
    \includegraphics[width=0.995\linewidth]{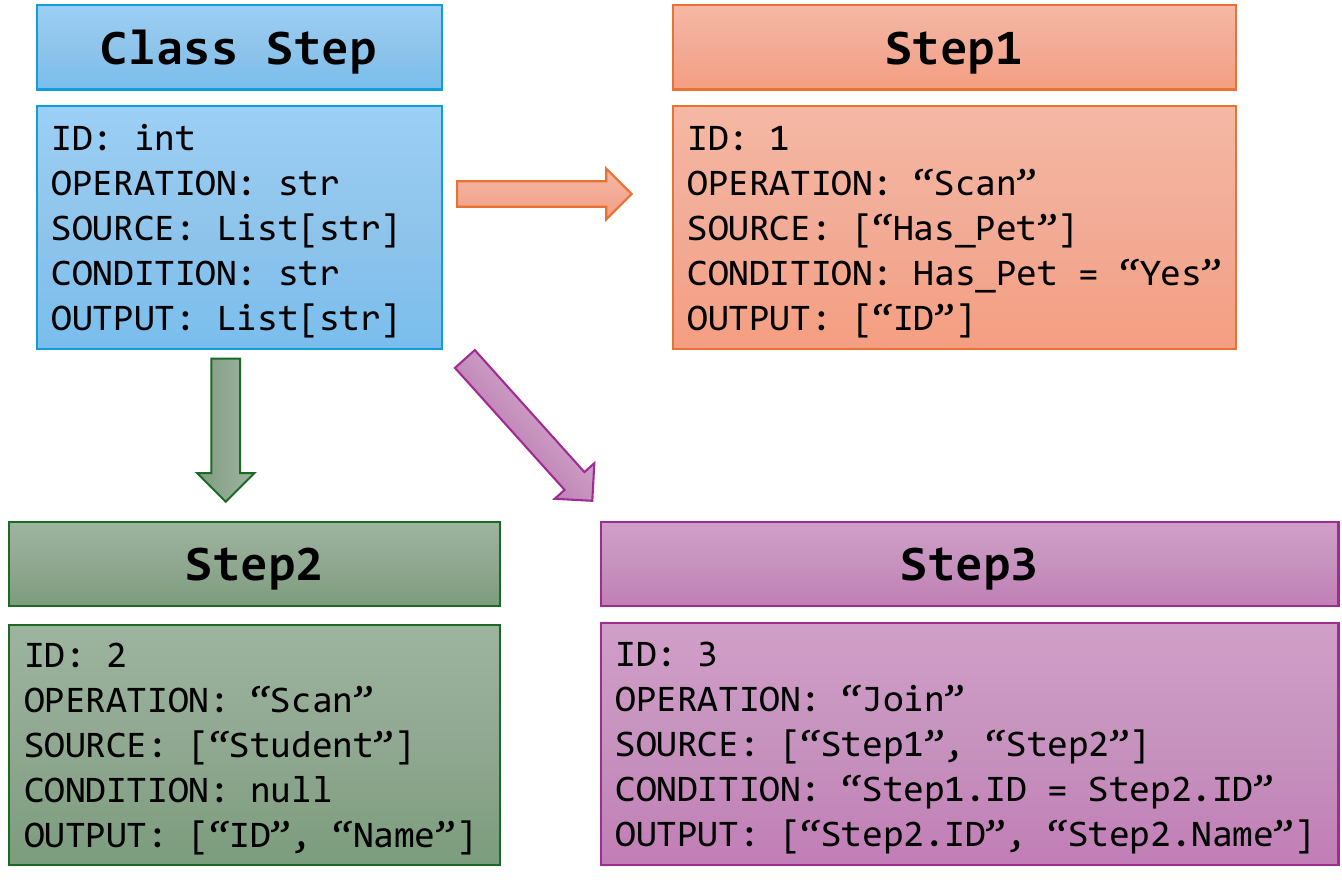}
    \caption{Class definition of our structured plan \ourplan alongside an example plan instance.}
    \label{fig:tabreason:plan_definition}
\end{figure}
\section{Problem Formulation.}

The task of query-focused table summarization involves generating a query-specific summary $s$ given a user query $q$ and a collection of relational tables $\mathcal{T} = \{ T_1, T_2, \ldots, T_n \}$. This process can be expressed as a function mapping: $(q, \mathcal{T}) \rightarrow s$, where $s$ is tailored to capture the information relevant to $q$.

\begin{figure*}[th]
    \centering
    \includegraphics[width=0.995\linewidth]{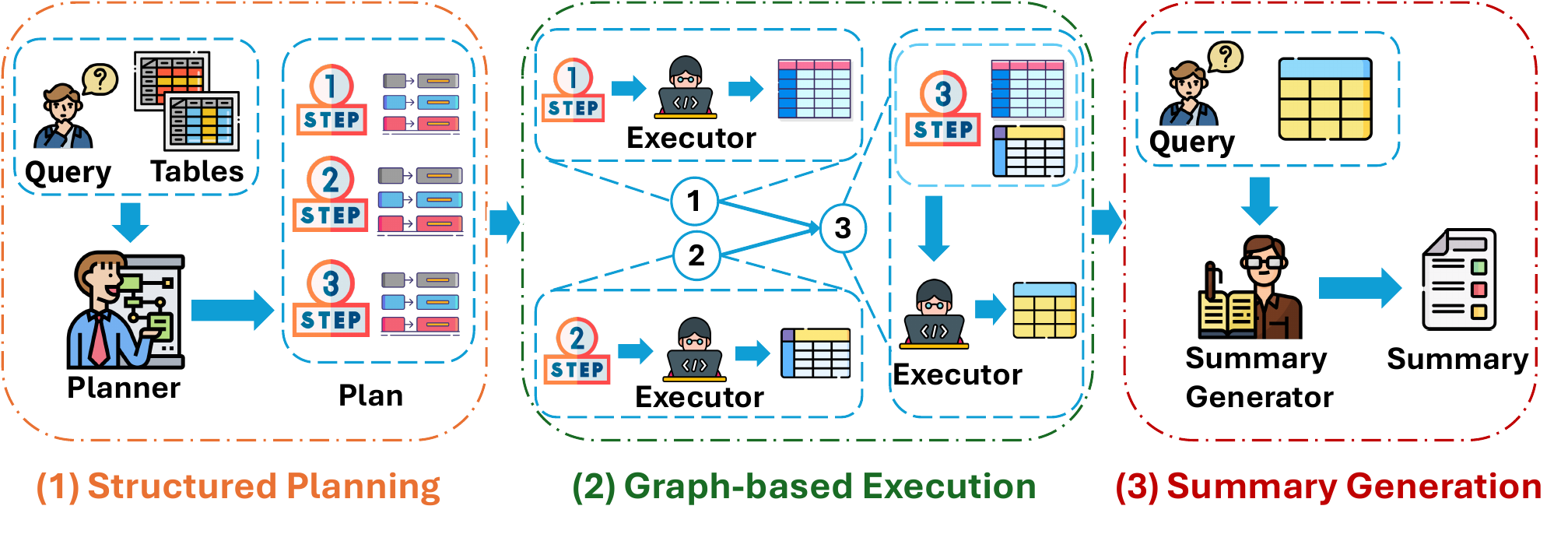}
    \caption{The overview of our framework \tabreasonours. First, the planner generates a plan during structured planning. Then, the executor performs graph-based execution by constructing a directed acyclic graph based on step dependencies and executes each planning step in parallel according to the graph structure. Finally, the summary generator synthesizes the last execution output into an informative summary tailored to the query.}
    \label{fig:tabreason:method_overview}
\end{figure*}
\section{Methodology}
\label{sec:tabreason:method}

In this section, we first provide an overview of our proposed framework, followed by a detailed explanation of each of its main components.

\subsection{Overview}

Our proposed framework follows a ``plan-then-execute'' paradigm~\cite{wang-etal-2023-plan}. It consists of three phases: \textit{Structure-based Planning}, \textit{Graph-based Execution} and \textit{Summary
Generation}, each corresponding to a key component: a planner, an executor, and a summary generator. An overview of the framework is illustrated in \autoref{fig:tabreason:method_overview}.
Specifically, during the planning phase, the planner generates multiple atomic and structured steps based on the given query $q$ and tables $\mathcal{T}$. In the execution phase, the executor first identifies dependencies among these steps and constructs a directed acyclic graph (DAG) from the sequential steps to determine the optimal execution order.
Next, the executor constructs and executes a corresponding SQL query for each step, leveraging the results of previous steps to produce intermediate execution tables. Finally, the summary generator synthesizes these execution tables into a coherent and concise textual summary.

\begin{table}[ht!]
  \centering
  \begin{tabular}{ll}
    \toprule
    \textbf{Operation} & \textbf{Functionality} \\
    \midrule
    \textbf{Scan} & Scan all table rows. \\
    \textbf{Aggregate} & Group and aggregate tuples. \\
    \textbf{Filter} & Remove non-matching tuples. \\
    \textbf{Sort} & Sort stream by expression. \\
    \textbf{TopSort} & Select top-K tuples. \\
    \textbf{Join} & Logically join two streams. \\
    \textbf{Except} & Compute set difference. \\
    \textbf{Intersect} & Compute set intersection. \\
    \textbf{Union} & Compute set union. \\
    \bottomrule
  \end{tabular}
  \caption{Definition of the nine operations used in the structured planning steps.}
  \label{tab:tablreason:table_operations}
\end{table}

\subsection{Structured Planning}
\label{subsec:tabreason:method:plan}

Structured planning aims to formalize the planning process with well-defined, structured planning steps.
Previous work on the plan-then-execute framework primarily relies on free-form natural language plans~\cite{zhao-etal-2024-tapera,nguyen2025interpretable}. However, such plans inherently contain linguistic ambiguities, making it challenging to automatically convert plan steps into executable programs during execution.
This work instead focuses on generating well-structured plans that align more effectively with the execution phase. 
Inspired by formalism in traditional multi-agent systems~\cite{dinverno1997formalism,hilaire2000formal,gruer2002formal}, we define a plan as a sequence of well-defined, structured steps. 

\paragraph{Plan Definition.}

Each step within the plan defines a table operation that transforms one or more source tables into a result table. The class definition for the step is shown in \autoref{fig:tabreason:plan_definition}. Specifically, we formulate a step as a standard class with attributes from software engineering: $S_i = \{\texttt{ID}, \texttt{OPERATION}, \texttt{SOURCE}, \texttt{CONDITION}, $
$\texttt{OUTPUT}\}$. Here, \texttt{ID} denotes the step identifier, and \texttt{OPERATION} specifies the table operation to be performed. Inspired by the query plan language (QPL)~\cite{eyal-etal-2023-semantic}, we consider nine table operations that cover the most common scenarios. The set of operations is shown in \autoref{tab:tablreason:table_operations}.
The \texttt{SOURCE} field specifies the source tables involved in the operation. These may include the original input tables or the intermediate result tables returned by previous steps. The \texttt{CONDITION} field defines any constraints or filters applied during the operation, such as row or column selections, or keys used for table joins. Finally, \texttt{OUTPUT} denotes the output columns of the resulting table. 

Consider the example plan consisting of three steps shown in \autoref{fig:tabreason:plan_definition}, the step with the field \texttt{ID} equal to $1$ indicates the first step. It aims to extract the column \textit{ID} from the table \textit{Has\_Pet} with the condition \textit{Has\_Pet = ``Yes''}.

\paragraph{Plan Generation.}

To generate a structured plan, we leverage the structured output capabilities of recent LLMs, such as \texttt{gpt-4o-mini}. Given a user query $q$ and input tables $\mathcal{T}$, we prompt the LLM to output a list of structured plans based on our predefined class schema.
Since LLMs process only textual inputs, we follow prior work~\cite{liu2022tapex,pal-etal-2023-multitabqa} and linearize each input table into a text format before feeding it to the model. For a table with $n$ columns and $m$ rows, the linearized version is:

\begin{align*}
& \textbf{table name:}\ name \qquad \ \textbf{col:} \ h_1 \mid \ldots \mid h_n \\
& \textbf{row\ 1:}\ c_{1,1} \mid \ldots \mid c_{1,n} \ldots \quad \textbf{row\ m:}\ c_{m,1} \mid \ldots \mid r_{m,n}.
\end{align*}

Previous approaches~\cite{wang2024cotable,zhao-etal-2024-tapera} typically linearize the entire table, which leads to a substantial increase in input length, especially when tables have hundreds of rows. In contrast, our experiments show that the table schema (i.e., column headers) plays the most critical role. Including only the top-$k$ rows ($k \ll m$) retains performance while significantly reducing token usage.

\subsection{Graph-based Execution}
\label{subsec:tabreason:method:execute}

The execution phase aims to obtain an intermediate execution table for each planning step $s_i$. We observe that the execution order of steps can often be parallelized, as different steps may depend on distinct input tables or on separate preceding steps.
Consider the example plan in \autoref{fig:tabreason:plan_definition}. We can observe that the first step is independent of the second step, as they rely on different source tables. Therefore, these two steps can be executed in parallel to improve execution efficiency.

\paragraph{Graph Construction.}

In general, each plan can be formulated as a directed acyclic graph (DAG), denoted as $G = (V, E)$, where $V = \{s_1, \ldots, s_n\}$ denotes the set of planning steps, and $E = \{e_{ij}\}$ represents the set of step dependencies. Specifically, an edge $e_{ij}$ exists if the output table from step $s_i$ is used as input to step $s_j$. This graph-based formulation allows the framework to explicitly capture inter-step dependencies and enables more efficient execution of the overall plan.

\paragraph{Program-Augmented Execution.}

Given a planning step, existing LLMs may struggle to generate well-formatted tables. To address this limitation, we enhance the LLM-based executor by incorporating an external code interpreter. 
Specifically, the executor leverages the code generation capabilities of the LLM to produce executable code. 
Following prior work~\cite{cheng2023binder,nguyen2025interpretable}, we prompt the LLM to generate a SQL query $q^{sql}_i$ for each step $s_i$. 
To improve the robustness and correctness of the generated SQL, we include several demonstrations in the prompt to support in-context learning~\cite{brown2020gpt3}. 
The code interpreter then executes the SQL query $q^{sql}_i$ to obtain the corresponding execution table.

\subsection{Summary Generation}
\label{subsec:tabreason:method:summ}

Once the execution process is complete, we leverage an LLM-based summary generator that takes the original query $q$ and the final table produced by the last step $s_n$ as input to generate a comprehensive summary. This summary is expected to integrate all query-relevant information derived from the input tables. 

\begin{table}[th]
\centering
\begin{adjustbox}{max width=0.5\textwidth} {
\begin{tabular}{lccc}
\toprule
\textbf{Property} & \textbf{\fetaqa} & \textbf{\qtsumm} & \textbf{\qfmts} \\
\midrule
Table Source & Wiki & Wiki & Database \\
Avg. \#Tables Per Example & 1.0 & 1.0 & 1.8 \\
Avg. Summary Length & 23.3 & 67.8 & 58.5 \\
Test Size & 2,003 & 1,078 & 608 \\
\bottomrule
\end{tabular}
}
\end{adjustbox}
\caption{Basic statistics of the three benchmarks used in our experiments.}
\label{tab:tabreason:data_stats}
\end{table}

\section{Experimental Setup}
\label{sec:tabreason:setup}

This section presents the research questions and the experimental setup used to evaluate our proposed framework. We describe the datasets, baseline models, evaluation metrics, and implementation details.

\subsection{Research Questions}

Our experiments are designed to address the following research questions (RQs):
\begin{itemize}
    \item \textbf{RQ1}: How does the proposed framework \tabreasonours perform compared to baseline models?
    \item \textbf{RQ2}: How does our \ourplan perform compared to natural language plan formats?
    \item \textbf{RQ3}: How efficient is our graph-based execution?
\end{itemize}

\subsection{Datasets}
\label{subsec:tabreason:datasets}

To evaluate the effectiveness of our approach, we utilize the three popular query-focused table summarization datasets, including \fetaqa~\cite{nan-etal-2022-fetaqa}, \qtsumm~\cite{zhao-etal-2023-qtsumm}, and \qfmts dataset~\cite{zhang2024qfmts}. \fetaqa and \qtsumm are designed for \textit{single-table} scenarios, both of which collect relational tables from Wikipedia. \fetaqa and \qtsumm provide manually annotated query-summary pairs aligned with the corresponding tables. In contrast, the \qfmts dataset is designed for \textit{multi-table} scenarios. It collects queries and relational tables from the database and includes annotated summaries tailored to the associated queries. Key statistics of three datasets are presented in \autoref{tab:tabreason:data_stats}.

\subsection{Baselines}
\label{subsec:tabreason:baselines}

We evaluate our approach by comparing it against two categories of baselines, including prompting-based and program-augmented models.

\paragraph{Prompt-based Models.}

These models aim to perform table reasoning by only relying on the internal reasoning capabilities of LLMs themselves.
\begin{itemize}
    \item \textbf{CoT}~\cite{wei2022cot} uses Chain-of-Thoughts (CoT) prompting to prompt LLMs to generate a summary given the query and relational tables.
    \item \textbf{\refactor}~\cite{zhao-etal-2023-qtsumm} adapts heuristic methods to extract query-relevant facts from given tables. Such facts are then concatenated with a given query into context for LLMs to generate a summary.
    \item \textbf{\directsumm}~\cite{zhang2024qfmts} tackles the task in a single phase, which extracts query-relevant facts from tables and synthesizes a comprehensive query-focused summary based on the facts jointly.
    \item \textbf{\reasonsumm}~\cite{zhang2024qfmts} addresses the task in two sequential phases: extracting query-relevant facts from tables and synthesizing a comprehensive query-focused summary based on the retrieved facts.
\end{itemize}

\paragraph{Program-Augmented Models.}

Unlike prompt-based models, program-augmented models utilize the code generation capabilities of LLMs to generate SQL queries and combine an external database engine to execute model-generated SQL queries, utilizing the execution results to augment the reasoning capabilities of LLMs.

\begin{itemize}
    \item \textbf{\binder}~\cite{cheng2023binder} converts the given NL query into either a Python program or a SQL query and executes the executable program to obtain the final answer. In our experiments, we use the version of \binder that generates SQL queries for a fair comparison.
    \item \textbf{\dater}~\cite{ye2023dater} leverages query decomposition to decompose a given query into sub-queries and convert each sub-query into a SQL query, and aggregate the execution results of the SQL query into a final summary.
    \item \textbf{\tapera}~\cite{zhao-etal-2024-tapera} generates natural language plans by performing query decomposition and uses an executable Python program to obtain the intermediate answer for each planning sub-query. Then it aggregates all intermediate query-answer pairs into a final summary.
\end{itemize}

\begin{table*}[th]

\centering
\begin{adjustbox}{max width=\textwidth} {
\begin{tabular}{l *{9}{c}}
    \toprule
        \multirow{2}{*}{\textbf{Method}}  & \multicolumn{3}{c}{\textbf{\fetaqa}}  & \multicolumn{3}{c}{\textbf{\qtsumm}} & \multicolumn{3}{c}{\textbf{\qfmts}} \\ 
        \cmidrule(lr){2-4} \cmidrule(lr){5-7} \cmidrule(lr){8-10}
       & \textbf{BLEU} & \textbf{ROUGE-L} & \textbf{METEOR} & \textbf{BLEU} & \textbf{ROUGE-L} & \textbf{METEOR} & \textbf{BLEU} & \textbf{ROUGE-L} & \textbf{METEOR}  \\
    \midrule
    \rowcolor{mygray!40}\multicolumn{10}{l}{\textbf{Prompt-based}}\\
         CoT           & 28.2  & 51.0 & 56.9  & 19.3  & 39.0  & 47.2 & 31.5 & 54.3 & 58.1 \\
         \refactor     & 26.2 & 53.6 & 57.2  & 19.9  & 39.5  & 48.8 & --- & --- & --- \\
         \directsumm    & 29.8 & 51.7 & 58.2  & 20.7  & 40.2  & 50.3 & 33.6 & 57.0 & 62.8  \\
         \reasonsumm  & \underline{31.7} & 52.6 & \underline{60.7}  & \textbf{21.8}  & \textbf{42.3}  & \textbf{51.5}  & 40.8 & 62.7 & 66.2 \\
    \midrule
    \rowcolor{mygray!40}\multicolumn{10}{l}{\textbf{Program-Augmented}}\\
        \binder   & 25.5 & 47.9 & 51.1   & 18.2  & 40.0  & 39.0 & \underline{42.5} & \underline{65.3} & \underline{70.7}  \\
        \dater    & 29.8  & \underline{54.0} & 59.4  & 16.6  & 35.2 & 35.5 & --- & --- & --- \\
        \tapera   & 29.5  & 53.4 & 58.2  & 14.6  & 33.0 & 33.2 & --- & --- & --- \\
       \textbf{\tabreasonours (Ours)}    & \textbf{33.8} & \textbf{55.7} & \textbf{62.3}  & \underline{20.9}  & \underline{41.3}  & \underline{47.7} & \textbf{45.7} & \textbf{68.3} & \textbf{73.4}  \\
    \bottomrule
\end{tabular}
} 
\end{adjustbox}
\caption{Automated evaluation of our approach and baseline models on the test sets of three benchmarks. \fetaqa\ and \qtsumm\ are single-table datasets, while \qfmts\ is a multi-table dataset. The best and second-best results are shown in \textbf{bold} and \underline{underline}, respectively.}
\label{tab:tabreason:main_results}
\end{table*}

\subsection{Automated Evaluation}
\label{subsec:tabreason:auto_eval}

Following \citet{zhao-etal-2024-tapera}, we evaluate the model's performance regarding the general quality of the generated summaries compared to the reference summaries, focusing on fluency and accuracy. We consider the following metrics:
\begin{itemize}
    \item \textbf{BLEU}~\cite{papineni-etal-2002-bleu} is a precision-oriented metric evaluating n-gram overlap between generated and reference summaries. We use SacreBLEU to compute BLEU scores.
    \item \textbf{ROUGE}~\cite{lin-hovy-2003-automatic} measures recall-oriented word overlap between generated and reference summaries. We report the F1 versions of ROUGE-L.
    \item \textbf{METEOR}~\cite{banerjee-lavie-2005-meteor} measures the unigram matching between the generated and reference summaries. 
\end{itemize}

\subsection{Implementation Details}
\label{subsec:tabreason:imple_details}

For the planner, executor, and summary generator in our framework, we adopt \texttt{gpt-4o-mini} as the backbone model due to its strong performance on table understanding tasks~\cite{nguyen2025interpretable}.
To enable effective in-context learning~\cite{brown2020gpt3}, we prepend the input prompt with $3$-shot demonstrations. We set the temperature, top-p, and maximum output tokens to $0.1$, $0.95$, and $400$, respectively. 
For the baseline models, we follow their original papers to set their hyperparameters.

\section{Results and Analyses}
\label{sec:tabreason:results}

\subsection{Main Results}
\label{subsec:tabreason:main_results}

To address \textbf{RQ1}, we report the evaluation results in Table~\ref{tab:tabreason:main_results}. The results show that \tabreasonours outperforms all baseline models on both the \fetaqa and \qfmts datasets, with especially strong results on \qfmts. 

The comparison between prompt-based and program-augmented methods shows that their performance depends on the dataset. 
On \qtsumm, prompt-based models like \reasonsumm generally perform better. In contrast, program-augmented methods such as \tabreasonours and \binder achieve higher scores on \qfmts. This difference mainly comes from the structure of the source tables. 
Tables in \qtsumm, collected from Wikipedia, often have multi-level and nested headers. 
SQL queries struggle with these complex structures, making it difficult for program-augmented methods to extract the correct information. 
In contrast, \qfmts mainly contains standard relational tables, which are easier for program-augmented methods to handle using SQL queries.

\begin{table}[t]
\centering
\begin{adjustbox}{max width=0.5\textwidth} {
\begin{tabular}{lcccc}
\toprule
\textbf{Plan Format} & \textbf{BL} & \textbf{R-L} & \textbf{ME} & \textbf{ESR} \\
\midrule
\qdmr & 42.3 & 65.1 & 69.4 & 94.4 \\
\qpl & 43.4 & 66.2 & 71.5 & 95.1 \\
\textbf{\ourplan (Ours)} & \textbf{45.7} & \textbf{68.3} & \textbf{73.4} & \textbf{98.2} \\
\bottomrule
\end{tabular}
}
\end{adjustbox}
\caption{Comparison of our \ourplan with  natural language plan formats on the \qfmts test set. BL, R-L, ME, and ESR denote BLEU, ROUGE-L, METEOR, and Execution Success Rate, respectively. ESR refers to the proportion of plan steps that can be successfully executed. The best are highlighted in \textbf{bold}.}
\label{tab:tabreason:ablation_plan_formats}
\end{table}
\begin{figure}[t]
    \centering
    \includegraphics[width=0.9\linewidth]{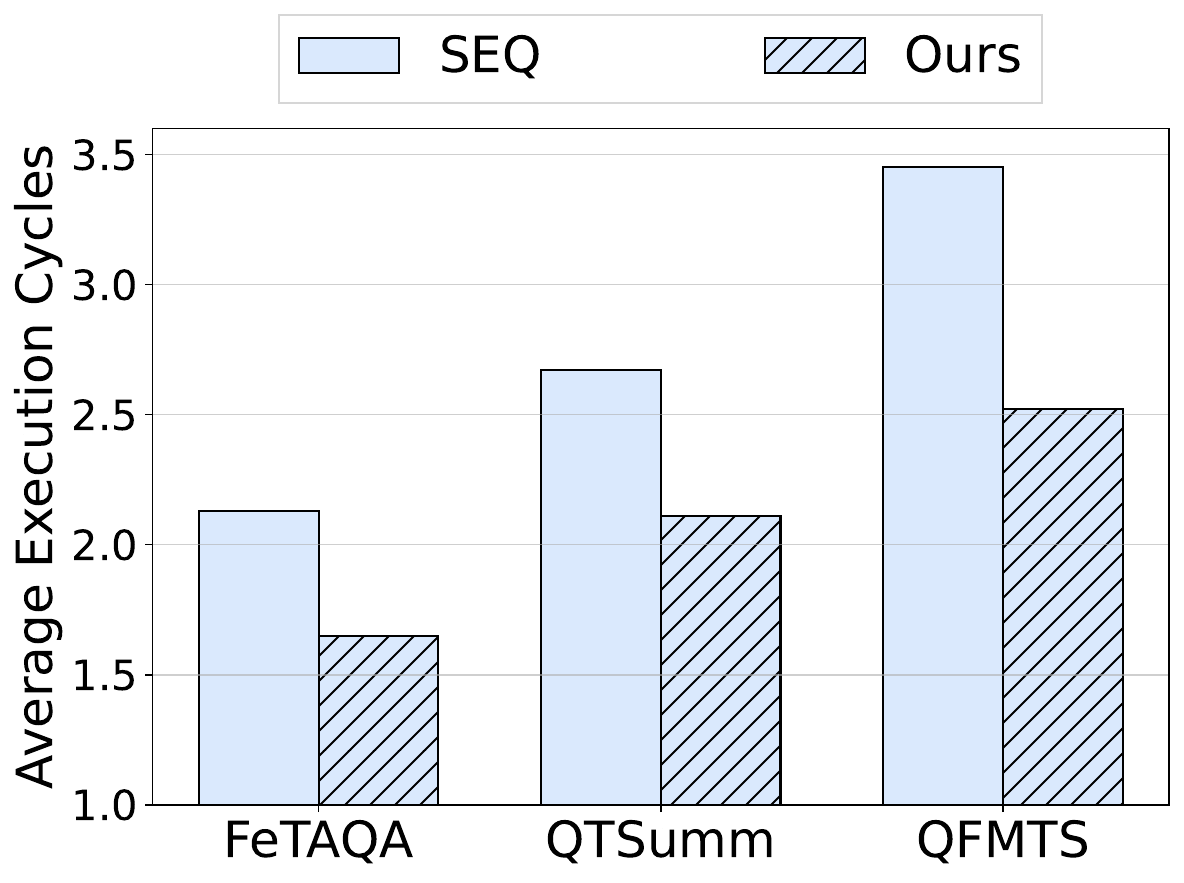}
    \caption{Comparison of the average number of execution cycles between previous sequential execution (SEQ) and our graph-based execution (Ours).}
    \label{fig:tabreason:execution_efficiency}
\end{figure}
\subsection{Effect of Plan Formats}
\label{subsec:tabreason:results:plan_format}

To address \textbf{RQ2}, we compare \ourplan with natural language plan formats, including QPL~\cite{eyal-etal-2023-semantic} and QDMR~\cite{wolfson-etal-2022-weakly}. To ensure a fair comparison, we adapt our planner to generate each plan format while keeping all other components unchanged. The results are shown in \autoref{tab:tabreason:ablation_plan_formats}.
From the table, we see that \ourplan consistently outperforms the NL plan formats across all metrics. For example, \ourplan achieves a METEOR (ME) score of 73.4, outperforming the next-best format, QPL.

Since program-augmented methods depend on the LLM's ability to turn each plan step into executable SQL, the accuracy of SQL generation is critical. However, LLMs may produce invalid SQL, which can cause run-time errors during execution. Thus, comparing the execution success rate across plan formats is important. As shown in \autoref{tab:tabreason:ablation_plan_formats}, \ourplan achieves a significantly higher execution success rate (ESR) of 98.2\%, compared to 95.1\% for QPL. This demonstrates the advantage of \ourplan in both SQL execution accuracy and reliability.

\begin{table*}[t]
\renewcommand\arraystretch{1.2}
\small
\centering
\resizebox{\linewidth}{!}{
\begin{tabular}{p{0.2\textwidth}<{\raggedright}p{0.8\textwidth}}
\toprule[1pt]
\textbf{\normalsize Challenge} & \textbf{\normalsize Representative Example} \\
\midrule[1pt]

\textbf{\normalsize Semantic Mismatch between Query and Table Schema} &

\textbf{Query:} What is the average duration of projects? 

\textbf{Tables:} 

\hspace*{1em} Table 1: Projects

\hspace*{2em} Table Schema: \textit{[ProjectID, ProjectName, StartDate, EndDate, Cost]}

\textbf{Natural Language Plan}:

\hspace*{1em} 1. \textit{Find \colorbox{myred}{duration} of each project.}

\hspace*{1em} 2. \textit{Calculate the average of these durations.}

\hspace*{1em} 3. \textit{Return the average duration for all projects.}

\textbf{Our \ourplan} $\{\texttt{ID; OPERATION; SOURCE; CONDITION; OUTPUT}\}$: 

\hspace*{1em} $\texttt{Step1} = \{\texttt{1}; \texttt{Scan}; \texttt{[Projects]}; \texttt{null}; 
\texttt{[ProjectID, StartDate, EndDate]}\}$ 

\hspace*{1em} $\texttt{Step2} = \{\texttt{2}; \texttt{Aggregate}; \texttt{[Step1]}; \texttt{null}; $

\qquad \qquad \qquad  $\texttt{[ProjectID, \colorbox{mygreen}{(EndDate - StartDate) as Duration]}}\}$

\hspace*{1em} $\texttt{Step3} = \{\texttt{3}; \texttt{Aggregate};\texttt{[Step2]}; \texttt{null}; 
\texttt{[AVG(Duration)]}\}$
\\
\midrule

\textbf{\normalsize Complex Step Dependencies in Multi-Table Scenarios} &

\textbf{Query:} List names and department locations of employees over 40 located in ``London''.

\textbf{Tables:} 

\hspace*{1em} Table 1: Employees

\hspace*{2em} Table Schema: \textit{[EmpID, EmpName, Age, DeptID]}

\hspace*{1em} Table 2: Departments

\hspace*{2em} Table Schema: \textit{[DeptCode, DeptName, Location]}

\textbf{Natural Language Plan}:

\hspace*{1em} 1. \textit{Get employees over 40 from the employees table.}

\hspace*{1em} 2. \textit{Find department locations in ``London'' from the departments table.}

\hspace*{1em} 3. \textit{\colorbox{myred}{Combine employee names and department locations}.}

\textbf{Our \ourplan} $\{\texttt{ID; OPERATION; SOURCE; CONDITION; OUTPUT}\}$:

\hspace*{1em} $\texttt{Step1} = \{\texttt{1}; \texttt{Scan}; \texttt{[Employees]}; \texttt{Age > 40}; 
\texttt{[EmpID, EmpName, DeptID]}\}$ 

\hspace*{1em} $\texttt{Step2} = \{\texttt{2}; \texttt{Scan}; \texttt{[Departments]}; \texttt{Location=``London''}]; \texttt{[DeptCode, Location}\}$

\hspace*{1em} $ \colorbox{mygreen}{\texttt{Step3} = \{\texttt{3}; \texttt{Join}; \texttt{[Step1, Step2]}; \texttt{Step1.DeptID = Step2.DeptCode};}$

\qquad \qquad \qquad  $\texttt{[Step1.EmpName, Step2.Location]}\}$
\\
\bottomrule

\end{tabular}
}
\caption{Qualitative comparison between the NL plan and our approach, \ourplan. We indicate two key challenges that lead to the failure of the NL plan. For simplicity, we flatten our class-based structured plan to improve readability. \colorbox{myred}{Red phrases} highlight key errors in the NL plan, while \colorbox{mygreen}{Green phrases} show how our plan addresses those errors.}
\label{tab:tabreason:qualitative_analysis}
\end{table*}
\subsection{Efficiency of Graph-based Execution}

To address \textbf{RQ3}, we measure execution efficiency using the number of execution cycles. 
In sequential execution, step dependencies are ignored, and steps are executed one after another. Therefore, the number of execution cycles is equal to the number of steps in the plan.
In contrast, graph-based execution accounts for step dependencies and enables parallel execution when possible. Here, the number of execution cycles corresponds to the length of the longest path in the DAG. 
For example, in the plan shown in \autoref{fig:tabreason:plan_definition}, there are three steps. Steps 1 and 2 can be executed in parallel during the first cycle, while Step 3, which depends on both, runs in the second cycle. Thus, the plan requires 2 execution cycles.

\autoref{fig:tabreason:execution_efficiency} compares the average number of execution cycles between sequential and graph-based execution. The results show that graph-based execution is more efficient, particularly on the \qfmts dataset. For example, it reduces the average number of execution cycles by about 28\%. This improvement is due to \qfmts involving multi-table inputs, where steps are more likely to operate on different input or intermediate tables. This allows for greater parallelism in SQL generation and execution. Consequently, the performance gap between the two execution methods is more pronounced for \qfmts than for \fetaqa and \qtsumm.

\subsection{Qualitative Analysis}

To highlight the strengths of our structured plan, \ourplan, we conduct a qualitative analysis comparing it to the NL plan, as shown in \autoref{tab:tabreason:qualitative_analysis}.
The analysis reveals two primary challenges that lead to the failure of the NL plan.

The first challenge is a semantic mismatch between the query and the table schema. This occurs due to unclear schema links between the query and the table.
In the corresponding example, the query uses the term \textit{duration}, which is not present in the table schema. The NL plan also includes an invalid step, ``\textit{Find duration of each project.}'', which leads to execution failure. In contrast, \ourplan restricts each step’s output to use only existing column names. It eliminates the ambiguity by calculating \textit{duration} using the \textit{StartDate} and \textit{EndDate} columns based on their meaning, effectively resolving the mismatch.

The second challenge is complex step dependencies, which is especially problematic when multiple input tables are involved.
In the corresponding example, the NL plan fails to connect the first two steps properly. As a result, the final step wrongly performs a union of their outputs, even though the query requires their intersection.
\ourplan addresses this by using the \texttt{SOURCE} field to explicitly link steps. For instance, \texttt{Step3} cites \texttt{Step1} and \texttt{Step2} as its sources, ensuring the correct interpretation of the query.
\section{Conclusion}
\label{sec:tabreason:conclu}

Query-focused table summarization aims to create concise, query-specific summaries from tabular data. While recent advancements leverage large language models (LLMs) for multi-step reasoning, a key challenge persists: natural language (NL) reasoning plans are inherently ambiguous, leading to unreliable execution and hindering scalability.
To address this, we propose \ourplan, a novel, structured plan format that formalizes a reasoning plan. Building on this, we introduce \tabreasonours, a framework that employs structured planning and graph-based execution for robust single- and multi-table summarization. Our evaluations show that \tabreasonours outperforms existing methods, validating the effectiveness of our structured approach in achieving more reliable and accurate table understanding.

\bibliography{main}

\end{document}